\documentclass[graybox]{svmult}

\usepackage{type1cm}        % activate if the above 3 fonts are
\usepackage{makeidx}         % allows index generation
\usepackage{graphicx} 
\usepackage{tabularx}% standard LaTeX graphics tool
\usepackage{multicol}        % used for the two-column index
\usepackage[bottom]{footmisc}% places footnotes at page bottom
\usepackage{url}
\usepackage{float}
\usepackage{amsmath}
\usepackage{xcolor}
\usepackage{soul}
\sethlcolor{yellow}
\usepackage[caption=false]{subfig}
\providecommand{\Description}[1]{}

\IfFileExists{newtxtext.sty}{%
  \usepackage{newtxtext}%
  \IfFileExists{newtxmath.sty}{\usepackage[varvw]{newtxmath}}{\usepackage{amssymb}}%
}{%
  \usepackage{mathptmx}%
  \usepackage{amssymb}%
}
\providecommand{\orcidID}[1]{} % Provide \orcidID 
\usepackage[hidelinks,hypertexnames=false,bookmarks=false]{hyperref}
\renewcommand{\orcidID}[1]{}
\makeindex             % used for the subject index
 \newcommand{\paperabstract}{Postoperative acute kidney injury (AKI) after major non-cardiac surgery carries
substantial morbidity, yet early intraoperative risk stratification remains
difficult. In this retrospective cohort study, we propose SynerT, a waveform-only hybrid
temporal backbone that combines a causal dilated TCN with a hierarchy of dilated
recurrent layers to encode early intraoperative physiologic trajectories for AKI
risk prediction.
Building on SynerT, we further design two model variants that extend the
backbone with structured clinical context: SynerT-MM, a late-fusion multimodal
extension that integrates hemodynamic burden summaries and preoperative
covariates, and SynerT-Stack, a leakage-safe stacked ensemble that combines
cross-validated predictions from SynerT-MM with strong tabular baselines at
the meta-learning stage.
All models are evaluated under a strict leakage-aware framework on VitalDB,
a high-fidelity perioperative database, with prediction restricted to information available within the first 60 intraoperative minutes. Among 2{,}413 waveform-usable cases (180 AKI-positive; 7.46\% prevalence), SynerT fell well
below strong structured-data baselines, demonstrating that waveform-only temporal
modeling is insufficient under strict early constraints. SynerT-MM recovered
discrimination by incorporating hemodynamic burden summaries and preoperative
covariates, and SynerT-Stack achieved the best overall performance across
AUROC, AUPRC, and F1-max. Cross-fitted Platt recalibration substantially corrected
calibration defects in both multimodal variants, and decision-curve analysis
confirmed the recalibrated stacked model delivered the strongest net clinical
benefit across low-to-intermediate thresholds. These findings indicate that
credible early perioperative AKI risk stratification requires structured
multimodal context, calibration-aware estimation, and transport robustness
evaluation rather than waveform modeling complexity alone.
}
\newcommand{\missingfigurebox}[1]{%
    \fbox{\parbox[c][0.22\textheight][c]{#1}{\centering Figure file unavailable in the current Overleaf project.}}%
}

\begin{document}

\def\orcidID#1{\unskip$^{[#1]}$}% restore ORCID display (svmult.cls disables it)
\title*{A Leakage-Aware Multimodal Evaluation Framework for Early Intraoperative Acute Kidney Injury Prediction}
\titlerunning{Multimodal Framework for Early AKI Prediction}
\author{Quang Minh Nguyen \orcidID{0009-0006-1135-5861} and \\
Duc Minh Le \orcidID{0009-0007-2269-5949} and \\
Ho Nhat Minh Nguyen \orcidID{0009-0008-4290-0759} and \\
Thuy Quynh Nguyen \orcidID{0009-0000-1241-5666} and \\
Trong Nghia Nguyen \orcidID{0000-0003-1888-0117}}
\authorrunning{Nguyen Q. Minh et al.}
\institute{Quang Minh Nguyen \at National Economics University, Hanoi, Vietnam, \email{11247324@st.neu.edu.vn}
\and Duc Minh Le \at National Economics University, Hanoi, Vietnam, \email{11247320@st.neu.edu.vn}
\and Ho Nhat Minh Nguyen \at National Economics University, Hanoi, Vietnam, \email{11247321@st.neu.edu.vn}
\and Thuy Quynh Nguyen \at National Economics University, Hanoi, Vietnam, \email{11247346@st.neu.edu.vn}
\and Trong Nghia Nguyen \at National Economics University, Hanoi, Vietnam, \email{nghiant@neu.edu.vn}}
\hypersetup{pdftitle={A Leakage-Aware Multimodal Evaluation Framework for Early Intraoperative Acute Kidney Injury Prediction},pdfauthor={Nguyen Quang Minh et al.}}
\maketitle
\begin{abstract}\paperabstract\end{abstract}


\section{Introduction}
\label{sec:introduction}

Postoperative acute kidney injury (PO-AKI) is a common complication after major
non-cardiac surgery, associated with substantial short-term morbidity and
long-term mortality \cite{kdigo2012,prowle2021poaki}. Clinically, the optimal AKI
prediction system must identify elevated risk early enough during the operation to
allow for proactive intraoperative and postoperative management. However, this is
technically challenging because perioperative kidney injury emerges from the
interaction between baseline patient vulnerability and evolving physiologic stress.
Recent VitalDB-based studies have demonstrated the feasibility of AKI prediction
using both interpretable ensembles and temporal deep-learning models
\cite{peng2021aki,park2025aki}, yet the evidence base remains methodologically
uneven, often lacking strict leakage control for early-prediction constraints and
rigorous ablation of waveform-only versus multimodal designs
\cite{zhang2022aki,kline2022multimodal}. Rather than claiming a fundamentally novel
architecture, this study contributes a disciplined, leakage-aware evaluation
framework for multimodal fusion under strict early-prediction constraints, with
SynerT as the proposed waveform-only temporal backbone at its core.

The contributions of this paper are as follows. First, we establish a strict
leakage-aware early-prediction framework that restricts dynamic inputs to the
first 60 intraoperative minutes, applies fold-specific preprocessing, and builds
stacked predictions only from out-of-fold outputs (i). Second, within this framework,
we show that waveform-only temporal modeling is insufficient for reliable early
PO-AKI prediction, as the original SynerT underperforms strong structured-data
baselines (ii). Third, we show that adding structured perioperative context and
leakage-safe stacking improves performance, with SynerT-MM and especially
SynerT-Stack yielding stronger discrimination and clinical utility after
cross-fitted Platt recalibration (iii). Finally, we provide a clinically oriented
evaluation that includes calibration diagnostics, decision-curve analysis, and a
cross-setting transport stress test on eICU-CRD Demo. We hypothesized that early
intraoperative waveform dynamics alone would be insufficient for clinically
meaningful PO-AKI prediction, and that integrating structured perioperative
context via leakage-safe ensembling is required for superior discrimination and
calibrated risk estimation (iv).

% ---------------------- RELATED WORK -----------------------------
\section{Related Work}
\label{sec:related_work}

The clinical framing of this study follows established PO-AKI literature.
KDIGO remains the standard basis for AKI definition \cite{kdigo2012}, although
perioperative studies must specify which criteria are operationalized. In the
present study, the endpoint is restricted to creatinine-based postoperative AKI
because urine-output data were unavailable. Prowle et al.~\cite{prowle2021poaki}
emphasized that PO-AKI after non-cardiac surgery reflects interactions among
baseline susceptibility, operative stress, and postoperative renal assessment,
which motivates combining structured renal-risk context with intraoperative
physiologic trajectories.

Among direct machine-learning comparators, Peng et al.~\cite{peng2021aki}
provide the closest AKI-specific VitalDB-linked reference for the waveform
baseline considered here. Their CISS 2021 study evaluated both interpretable
ensemble learning on structured perioperative variables and an
attention-weighted CNN--LSTM for temporal perioperative signals, showing that
intraoperative sequence modeling is clinically informative while strong tabular
models remain difficult to outperform. Park et al.~\cite{park2025aki} likewise
showed that preoperative context and intraoperative physiologic signals are
complementary rather than interchangeable, while Lee et al.~\cite{lee2024open}
demonstrated that boosting-based tabular models remain highly competitive when
perioperative summary variables are available. Building on these studies, the
present work asks how much performance can be recovered when structured context
and leakage-safe ensembling are added to waveform-first modeling.

The rationale for multimodal integration is also consistent with broader
precision-health literature. Kline et al.~\cite{kline2022multimodal} reviewed
multimodal machine learning in precision health and found that data fusion often
improves predictive performance, but also noted limited evidence on deployment,
subpopulation robustness, and clinically meaningful evaluation. Similar concerns
have been raised for perioperative AKI prediction \cite{zhang2022aki}. Because
AKI prediction models may function as clinical risk scores rather than only
ranking systems, calibration and clinical utility are central evaluation
components. Prior work has emphasized the importance of calibration assessment,
transparent reporting, and decision-curve analysis for clinically meaningful
prediction models \cite{vancalster2019,collins2024tripodai,vickers2006dca,vickers2021dca}.

Taken together, prior studies support the feasibility of perioperative AKI
prediction and the complementary value of multimodal perioperative information,
while also showing that strong tabular baselines are difficult to surpass.
However, the literature still incompletely characterizes how much early
predictive signal arises from waveform trajectories alone versus structured
renal-risk context under strict leakage-control settings. This gap motivates the
present ablation-focused evaluation of waveform-only modeling, multimodal fusion,
leakage-safe stacking, calibration, and transport robustness within a unified
early-prediction framework.

% ------------------------- METHODOLOGY ------------------------------
\section{Methodology}
\label{sec:methodology}

\begin{figure}[H]
\centering
\Description{Overview of the SynerT-MM pipeline with leakage-aware stacking:
temporal waveform preprocessing, TCN/dilated-RNN encoding, structured feature
embedding, late fusion, and meta-learner ensemble for final AKI risk.}
\IfFileExists{figure/synert_pipeline.png}{%
\includegraphics[width=0.92\textwidth]{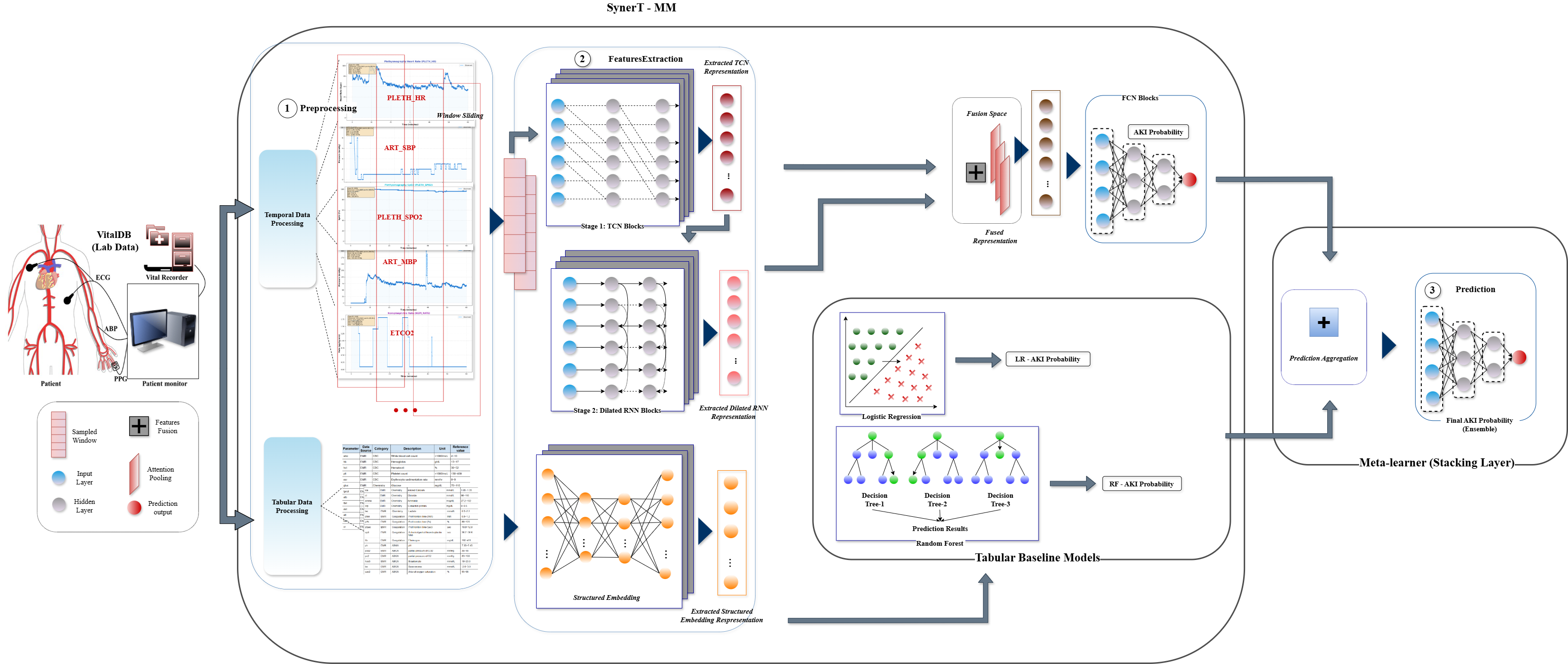}
}{%
\missingfigurebox{0.92\textwidth}%
}
\caption{Overview of the SynerT-MM pipeline with leakage-aware stacking:
temporal waveform preprocessing, temporal encoding, structured feature embedding,
late fusion, and meta-learner ensemble for final AKI risk.
The original SynerT backbone is the waveform-only temporal branch, while SynerT-MM
and SynerT-Stack extend that backbone with structured context and leakage-safe
ensembling, respectively.}
\label{fig:synert_flowchart}
\end{figure}

\subsection{Notation}
\label{subsec:notation}

In the Methods, $i$ indexes surgical cases and $t$ indexes seconds within the
early intraoperative window.
$T_i$ denotes the valid sequence length for case $i$;
$\mathbf{x}_{i,t} \in \mathbb{R}^{C}$ the resampled physiologic value vector;
$\mathbf{m}_{i,t} \in \{0,1\}^{C}$ the corresponding observation-mask vector;
$\mathbf{z}_{i,t} \in \mathbb{R}^{2C}$ the concatenated value-indicator input;
$\mathbf{Z}_i \in \mathbb{R}^{T_i \times 2C}$ the full temporal sequence;
$\mathbf{s}_i \in \mathbb{R}^{d_s}$ the structured static covariate vector;
and $a_{i,t}$ the intraoperative mean arterial pressure (MAP) series.
We write $\tau \in \{65, 60\}$ for the MAP hypotension thresholds used for
burden summaries.
The target is a binary postoperative AKI label $y_i \in \{0,1\}$, and model
outputs are scalar probabilities $\hat{p}_i \in (0,1)$.

\subsection{Data Source, Cohort, and Outcome Definition}
\label{subsec:data_cohort}

We used the VitalDB perioperative database, which contains synchronized waveform,
numeric, clinical, and laboratory records from surgical patients
\cite{lee2022vitaldb}.
The primary creatinine-labeled cohort contained 3{,}242 cases, of which 2{,}413
passed waveform-availability and quality screening for waveform-based development
and evaluation; 180 were AKI-positive (7.46\%).

The target was creatinine-based postoperative AKI aligned with the creatinine
component of KDIGO \cite{kdigo2012}.
For case $i$, let $c_i^{\text{base}}$ denote the most recent preoperative
creatinine within 30 days before surgery, and let $c_i^{\text{post}}$ denote the
maximum postoperative creatinine within 7 days.
The binary endpoint was defined as
\begin{equation}
  y_i = \mathbf{1}\!\left[
    c_i^{\text{post}} \ge 1.5\,c_i^{\text{base}}
    \;\lor\;
    \bigl(c_i^{\text{post}} - c_i^{\text{base}}\bigr) \ge 0.3~\text{mg/dL}
  \right].
  \label{eq:aki_label}
\end{equation}

% =========================================================
\subsection{Input Representation}
\label{subsec:input}
% =========================================================

\subsubsection{Dynamic Physiologic Signals}
\label{subsubsec:dynamic_signals}

Dynamic inputs were constructed from the first 60 intraoperative minutes and
resampled to a 1~Hz grid, yielding at most $T_i \le 3{,}600$ time steps per case.
For each case $i$ and second $t$, $\mathbf{x}_{i,t} \in \mathbb{R}^{C}$ denotes
the resampled physiologic values across $C = 7$ monitoring channels, and
$\mathbf{m}_{i,t} \in \{0,1\}^{C}$ denotes the matched observation indicators.
The retained channels were invasive mean arterial pressure,
plethysmographic pulse rate, peripheral oxygen saturation, invasive systolic
arterial pressure, invasive diastolic arterial pressure, heart rate, and
end-tidal carbon dioxide.

Each time step is represented by value-indicator concatenation,
\begin{equation}
  \mathbf{z}_{i,t} = \bigl[\mathbf{x}_{i,t};\,\mathbf{m}_{i,t}\bigr] \in \mathbb{R}^{2C}, \qquad t = 1,\ldots,T_i,
  \label{eq:joint_encoding}
\end{equation}

and the full temporal input sequence for case $i$ is
\begin{equation}
  \mathbf{Z}_i = \bigl(\mathbf{z}_{i,1}, \ldots, \mathbf{z}_{i,T_i}\bigr) \in \mathbb{R}^{T_i \times 2C}, \qquad 2C = 14.
  \label{eq:temporal_sequence}
\end{equation}

This representation preserves missingness explicitly rather than masking it
through imputation alone.
Linear interpolation was applied only within each channel's observed support;
outside that support, the value channel was zeroed and the mask channel recorded
absence of measurement.
Value channels were normalized using per-fold scalers fitted on training-fold
data only to prevent leakage, using standard $z$-score normalization for most
channels and robust scaling for the skewed vasopressor infusion rate.

\subsubsection{Structured Context and Hemodynamic Burden}
\label{subsubsec:structured_context}

To represent baseline renal-risk context, the multimodal models additionally
used a static covariate vector $\mathbf{s}_i \in \mathbb{R}^{d_s}$ containing
preoperative demographic, laboratory, and comorbidity information.
We also derived hemodynamic burden summaries from intraoperative MAP $a_{i,t}$
at the common hypotension thresholds $\tau \in \{65,60\}$~mmHg.
On the 1~Hz grid, three burden statistics were computed for each threshold $\tau$:

\begin{equation}
  B_{\tau,i}^{\text{time}} = \sum_{t=1}^{T_i} \mathbf{1}[a_{i,t} < \tau],
  \qquad
  B_{\tau,i}^{\text{frac}} = \frac{1}{T_i}\sum_{t=1}^{T_i} \mathbf{1}[a_{i,t} < \tau],
  \label{eq:burden_time_frac}
\end{equation}
\begin{equation}
  B_{\tau,i}^{\text{auc}} = \sum_{t=1}^{T_i} (\tau - a_{i,t})_{+},
  \label{eq:burden_auc}
\end{equation}
where $B_{\tau,i}^{\text{time}}$ is the total seconds below threshold $\tau$,
$B_{\tau,i}^{\text{frac}}$ is the corresponding fraction of the observation
window, and $B_{\tau,i}^{\text{auc}}$ is the area under the MAP-deficit curve.
These features were supplemented by episode counts, longest hypotensive run
duration, and short-term MAP variability statistics.
Burden summaries are interpreted here as clinically motivated exposure
descriptors rather than causal threshold claims.

% =========================================================
\subsection{SynerT: Waveform-Only Temporal Backbone}
\label{subsec:synert_backbone}
% =========================================================

We first describe SynerT, the original waveform-only backbone, because the
remaining model families either extend this encoder or act as stand-alone
tabular baselines.
SynerT takes as its sole input the value-indicator sequence
\begin{equation}
  \underbrace{\mathbf{Z}_i \in \mathbb{R}^{T_i \times 2C}}_{\text{input } X_i}
  \xrightarrow{\text{SynerT}}
  \underbrace{\hat{p}_i^{\text{SynerT}} \in (0,1)}_{\text{output } Y_i}
  \label{eq:synert_io}
\end{equation}
with no access to static clinical covariates or hemodynamic burden summaries.
The backbone consists of four stages.

\subsubsection{Stage 1: Channel Projection}
\label{subsubsec:stage_projection}

The input sequence is transposed from time-major to channel-major format,
$\mathbf{Z}_i \in \mathbb{R}^{T_i \times 2C} \to \mathbf{Z}_i^{\top}
\in \mathbb{R}^{2C \times T_i}$,
and projected into a $d_h$-dimensional latent space via a pointwise
($1 \times 1$) convolution:
\begin{equation}
  \mathbf{H}_i^{(0)}
  = \operatorname{Conv1D}_{1\times 1}\!\bigl(\mathbf{Z}_i^{\top}\bigr)
  \in \mathbb{R}^{d_h \times T_i}.
  \label{eq:input_projection}
\end{equation}
This step performs cross-channel mixing before temporal modeling in a uniform
$d_h$-dimensional feature space.

\subsubsection{Stage 2: Causal Dilated TCN}
\label{subsubsec:stage_tcn}

$\mathbf{H}_i^{(0)}$ is passed through $L_{\text{TCN}}$ residual TCN blocks with
exponentially increasing dilation.
Each block applies two causal 1-D convolutions (kernel size $k$, dilation
$d_\ell = 2^{\ell-1}$), each followed by ReLU and dropout, with a residual skip
connection:
\begin{equation}
  \mathbf{H}_i^{(\ell)}
  = \mathbf{H}_i^{(\ell-1)}
  + \phi_{\text{TCN}}^{(\ell)}\!\bigl(\mathbf{H}_i^{(\ell-1)}\bigr),
  \qquad \ell = 1,\ldots,L_{\text{TCN}},
  \label{eq:tcn_block}
\end{equation}
where
\begin{equation}
  \phi_{\text{TCN}}^{(\ell)}(\cdot)
  =
  \operatorname{Drop}\!\Bigl(
    \operatorname{ReLU}\!\Bigl(
      \operatorname{CausalConv}_{k,\,d_\ell}\!\Bigl(
        \operatorname{Drop}\!\bigl(
          \operatorname{ReLU}\!\bigl(
            \operatorname{CausalConv}_{k,\,d_\ell}(\cdot)
          \bigr)
        \bigr)
      \Bigr)
    \Bigr)
  \Bigr).
  \label{eq:tcn_block_expand}
\end{equation}
Causality is enforced by left-padding each convolution by $(k-1)d_\ell$
positions and trimming the future-facing outputs, so that
$\mathbf{H}_{i,t}^{(\ell)}$ depends only on
$\mathbf{z}_{i,1},\ldots,\mathbf{z}_{i,t}$.
With doubling dilation across levels, the TCN captures multi-scale temporal
patterns within the 60-minute observation window.

\subsubsection{Stage 3: Dilated RNN Hierarchy and Fusion}
\label{subsubsec:stage_rnn_fusion}

The TCN output $\mathbf{H}_i^{(L_{\text{TCN}})}$ is transposed back to
time-major format and processed by a hierarchy of $L_{\text{RNN}}$ dilated
recurrent layers $\phi_{\text{RNN}}^{(1)}, \ldots, \phi_{\text{RNN}}^{(L_{\text{RNN}})}$.
Each layer operates on a temporally sub-sampled input, expanding effective
sequential context without proportionally increasing recurrent steps.
The outputs of the TCN and RNN branches are then fused through a learned gating
mechanism $\phi_{\text{fuse}}$:
\begin{equation}
  \tilde{\mathbf{H}}_i
  = \phi_{\text{fuse}}\!\Bigl(
      \bigl[
        \mathbf{H}_i^{(L_{\text{TCN}})},\;
        \phi_{\text{RNN}}^{(1)}(\mathbf{Z}_i),\;
        \ldots,\;
        \phi_{\text{RNN}}^{(L_{\text{RNN}})}(\mathbf{Z}_i)
      \bigr]
    \Bigr)
  \in \mathbb{R}^{T_i \times d_h},
  \label{eq:fusion}
\end{equation}
This fused representation combines local multi-scale pattern extraction from the
TCN with longer-range sequential dependencies from the recurrent hierarchy.

\subsubsection{Stage 4: Masked Temporal Pooling and Classification Head}
\label{subsubsec:stage_pooling_head}

$\tilde{\mathbf{H}}_i$ is reduced to a fixed-length case embedding by masked mean
pooling over the valid temporal extent $T_i$, excluding zero-padded positions:
\begin{equation}
  \mathbf{h}_i^{\text{temp}}
  = \frac{1}{T_i} \sum_{t=1}^{T_i} \tilde{\mathbf{H}}_{i,t}
  \;\in\; \mathbb{R}^{d_h}.
  \label{eq:masked_pool}
\end{equation}
Dividing by $T_i$ rather than padded batch length avoids penalizing cases with
shorter valid observation windows.
The pooled embedding is then passed through dropout and a linear projection to
produce the AKI risk probability:
\begin{equation}
  \hat{p}_i^{\text{SynerT}}
  = \sigma\!\bigl(\mathbf{w}^{\top}\mathbf{h}_i^{\text{temp}} + b\bigr)
  \in (0,1),
  \label{eq:synert_head}
\end{equation}
where $\sigma(\cdot)$ is the sigmoid function.
In summary, SynerT maps
$\mathbf{Z}_i \in \mathbb{R}^{T_i \times 2C}
\;\longrightarrow\;
\hat{p}_i^{\text{SynerT}} \in (0,1)$
using waveform data alone.

% =========================================================
\subsection{Model Families and Variants}
\label{subsec:model_variants}
% =========================================================

The analysis centers on three model families.
The first is the original waveform-only SynerT backbone, which tests whether
early physiologic trajectories are sufficient on their own.
The second is a late-fusion multimodal extension (SynerT-MM) that adds
structured clinical covariates and hemodynamic burden summaries.
The third is a leakage-safe stacked ensemble (SynerT-Stack) that combines
cross-validated predictions from SynerT-MM with strong tabular baselines at the
meta-learning stage.

\subsubsection{SynerT-MM}
\label{subsubsec:synert_mm}

The input to SynerT-MM is the pair
\begin{equation}
  X_i^{\text{MM}} := \bigl(\mathbf{Z}_i,\; \mathbf{s}_i\bigr),
  \label{eq:mm_input}
\end{equation}
where $\mathbf{Z}_i \in \mathbb{R}^{T_i \times 2C}$ is the temporal waveform
sequence and $\mathbf{s}_i \in \mathbb{R}^{d_s}$ is the structured static
feature vector.
The temporal branch encodes $\mathbf{Z}_i$ through the SynerT backbone,
producing $\mathbf{h}_i^{\text{temp}} \in \mathbb{R}^{d_h}$ as in
Equations~\eqref{eq:input_projection}--\eqref{eq:masked_pool}.
The structured branch independently encodes $\mathbf{s}_i$ via a learned
projection $\phi_{\text{stat}}: \mathbb{R}^{d_s} \to \mathbb{R}^{d_h}$.
Late fusion is performed by concatenation after separate encoding,
\begin{equation}
  \mathbf{h}_i^{\text{MM}}
  = \bigl[\mathbf{h}_i^{\text{temp}};\; \phi_{\text{stat}}(\mathbf{s}_i)\bigr]
  \in \mathbb{R}^{2d_h},
  \label{eq:mm_concat}
\end{equation}
and the combined representation is passed through a classification head:
\begin{equation}
  Y_i^{\text{MM}} :=
  \hat{p}_i^{\text{MM}}
  = \sigma\!\bigl(\psi_{\text{MM}}(\mathbf{h}_i^{\text{MM}})\bigr)
  \in (0,1),
  \label{eq:mm_output}
\end{equation}
where $\psi_{\text{MM}}: \mathbb{R}^{2d_h} \to \mathbb{R}$ is the fusion
classification head.
Late fusion lets the temporal and structured branches be encoded separately
before prediction.

\subsubsection{SynerT-Stack}
\label{subsubsec:synert_stack}

SynerT-Stack does not operate on raw temporal sequences directly.
Its input consists of out-of-fold base-model probability estimates together with
leakage-safe static features:
\begin{equation}
  X_i^{\text{Stack}} :=
  \Bigl(
    \hat{p}_i^{\text{MM,OOF}},\;
    \hat{p}_i^{\text{RF,OOF}},\;
    \hat{p}_i^{\text{LR,OOF}},\;
    \mathbf{s}_i
  \Bigr),
  \label{eq:stack_input}
\end{equation}
where $\hat{p}_i^{\text{MM,OOF}}$, $\hat{p}_i^{\text{RF,OOF}}$, and
$\hat{p}_i^{\text{LR,OOF}}$ are out-of-fold probability estimates from
SynerT-MM, random forest, and logistic regression, respectively.
A logistic meta-learner is fit as
\begin{equation}
  Y_i^{\text{Stack}} :=
  \hat{p}_i^{\text{Stack}}
  = \sigma\!\Bigl(
      \beta_0
      + \beta_1 \hat{p}_i^{\text{MM,OOF}}
      + \beta_2 \hat{p}_i^{\text{RF,OOF}}
      + \beta_3 \hat{p}_i^{\text{LR,OOF}}
      + \boldsymbol{\gamma}^{\top} \mathbf{s}_i
    \Bigr)
  \in (0,1),
  \label{eq:stack_model}
\end{equation}
where $\beta_0 \in \mathbb{R}$ is the meta-learner intercept,
$\beta_1, \beta_2, \beta_3 \in \mathbb{R}$ are the base-model coefficients, and
$\boldsymbol{\gamma} \in \mathbb{R}^{d_s}$ is the coefficient vector for the
static safe features retained at the meta-learning stage.
Because the meta-learner is trained only on out-of-fold predictions,
SynerT-Stack avoids meta-level information leakage by construction.

The main text retains only the strongest or literature-grounded comparators:
the Peng-style CNN-LSTM baseline and high-capacity tree ensembles
(random forest, extra trees, and CatBoost-style boosting), aligned with recent
perioperative AKI baselines \cite{peng2021aki,park2025aki,lee2024open}.
\hyperref[tab:model_inputs]{Table~\ref*{tab:model_inputs}} summarizes the
information sources available to the main model families.

\begin{table}[t]
\caption{Information sources used by the main model families. Bold entries highlight
  the components that distinguish the multimodal and stacked variants from the
  original waveform-only model.}
\label{tab:model_inputs}
\centering
\small
\begin{tabularx}{\linewidth}{p{2.8cm}*{5}{>{\centering\arraybackslash}X}}
\hline\noalign{\smallskip}
Model family &
Temporal physiology &
Observation indicators &
Structured covariates &
Burden summaries &
Cross-validated model scores \\
\noalign{\smallskip}\hline\noalign{\smallskip}
Original SynerT    & Yes & Yes & No           & No           & No           \\
RF baseline        & No  & No  & \textbf{Yes} & \textbf{Yes} & No           \\
SynerT-MM          & Yes & Yes & \textbf{Yes} & \textbf{Yes} & No           \\
SynerT-Stack       & No  & No  & \textbf{Yes} & \textbf{Yes} & \textbf{Yes} \\
\noalign{\smallskip}\hline\noalign{\smallskip}
\end{tabularx}
\end{table}

\subsection{Evaluation Design and Leakage Control}
\label{subsec:evaluation_design}

All experiments used fold-specific preprocessing with scalers fitted on training
folds only, and stacked predictions were constructed exclusively from
out-of-fold base-model outputs to prevent meta-level leakage.
Performance was evaluated by five-fold cross-validation using AUROC and AUPRC
as the primary discrimination metrics, with particular emphasis on AUPRC given
the 7.46\% event prevalence.
Additional evaluation included operating-point metrics at 95\% specificity,
pooled out-of-fold calibration diagnostics (Brier score, ECE, reliability plots,
calibration intercept and slope), and decision-curve analysis over threshold
probabilities from 0.02 to 0.25
\cite{vancalster2019,collins2024tripodai,vickers2006dca,vickers2021dca}.
For SynerT-MM and SynerT-Stack, absolute-risk calibration was further improved
via leakage-safe cross-fitted Platt recalibration.
The eICU-CRD Demo cohort served as a transport stress test, scored without
refitting \cite{riley2024external,riley2021valsize}.

\section{Experiments and Results}
\label{sec:results}

\subsection{Main Model Comparison}
\label{subsec:main_comparison}

\hyperref[tab:main_results]{Table~\ref*{tab:main_results}} shows a clear
progression across model families. The original SynerT remained well below the
strongest structured-data baselines, confirming that waveform-only temporal
modeling is insufficient under strict early-prediction constraints.
Adding structured clinical context and hemodynamic burden summaries improved
discrimination in SynerT-MM, though the gain over random forest was directional
rather than statistically significant at the five-fold level.
SynerT-Stack achieved the highest AUROC, AUPRC, and F1-max across all models,
making leakage-safe stacking the clearest supported improvement in the main
comparison.
Secondary tabular baselines and additional waveform backbones did not surpass SynerT-MM or SynerT-Stack, reinforcing
that the core gain came from multimodal context and stacking rather than broader
backbone search.

\begin{table}[H]
\caption{Main five-fold performance results centered on the strongest and literature-grounded comparators. Values are mean $\pm$ standard deviation across validation folds. Boldface marks the best value in each column.}
\label{tab:main_results}
\centering
\small
\begin{tabular}{p{3.6cm}ccc}
\hline\noalign{\smallskip}
Model & AUROC & AUPRC & F1-max \\
\noalign{\smallskip}\hline\noalign{\smallskip}
CNN-LSTM & 0.600 $\pm$ 0.071 & 0.166 $\pm$ 0.047 & 0.250 $\pm$ 0.058 \\
Original SynerT & 0.648 $\pm$ 0.051 & 0.172 $\pm$ 0.026 & 0.245 $\pm$ 0.042 \\
Random-forest baseline & 0.729 $\pm$ 0.048 & 0.206 $\pm$ 0.045 & 0.311 $\pm$ 0.041 \\
Extra-trees baseline & 0.715 $\pm$ 0.044 & 0.197 $\pm$ 0.050 & 0.307 $\pm$ 0.045 \\
CatBoost baseline & 0.708 $\pm$ 0.026 & 0.209 $\pm$ 0.049 & 0.315 $\pm$ 0.051 \\
SynerT-MM & 0.734 $\pm$ 0.046 & 0.236 $\pm$ 0.041 & 0.349 $\pm$ 0.050 \\
SynerT-Stack & \textbf{0.773} $\pm$ \textbf{0.032} & \textbf{0.252} $\pm$ \textbf{0.031} & \textbf{0.367} $\pm$ \textbf{0.034} \\
\noalign{\smallskip}\hline\noalign{\smallskip}
\end{tabular}
\end{table}

\subsection{Cross-Setting Transport Stress Test on eICU-CRD Demo}
\label{subsec:eicu_external}

A preliminary cross-setting transport stress test on eICU-CRD Demo suggested substantial degradation for most standalone models under setting shift. However, SynerT-Stack retained the strongest discrimination (AUROC 0.831 $\pm$ 0.026; AUPRC 0.555 $\pm$ 0.177). Because the cohort was small and not a matched intraoperative surgical population, these findings should be interpreted only as a proxy transport result rather than definitive external validation.

\begin{figure}[H]
\centering
\Description{Two-panel bar chart comparing internal VitalDB and eICU Demo AUROC and AUPRC for SynerT, Random Forest, SynerT-MM, and SynerT-Stack.}
\IfFileExists{figure/revision_transport_shift.png}{%
\includegraphics[width=0.8\textwidth]{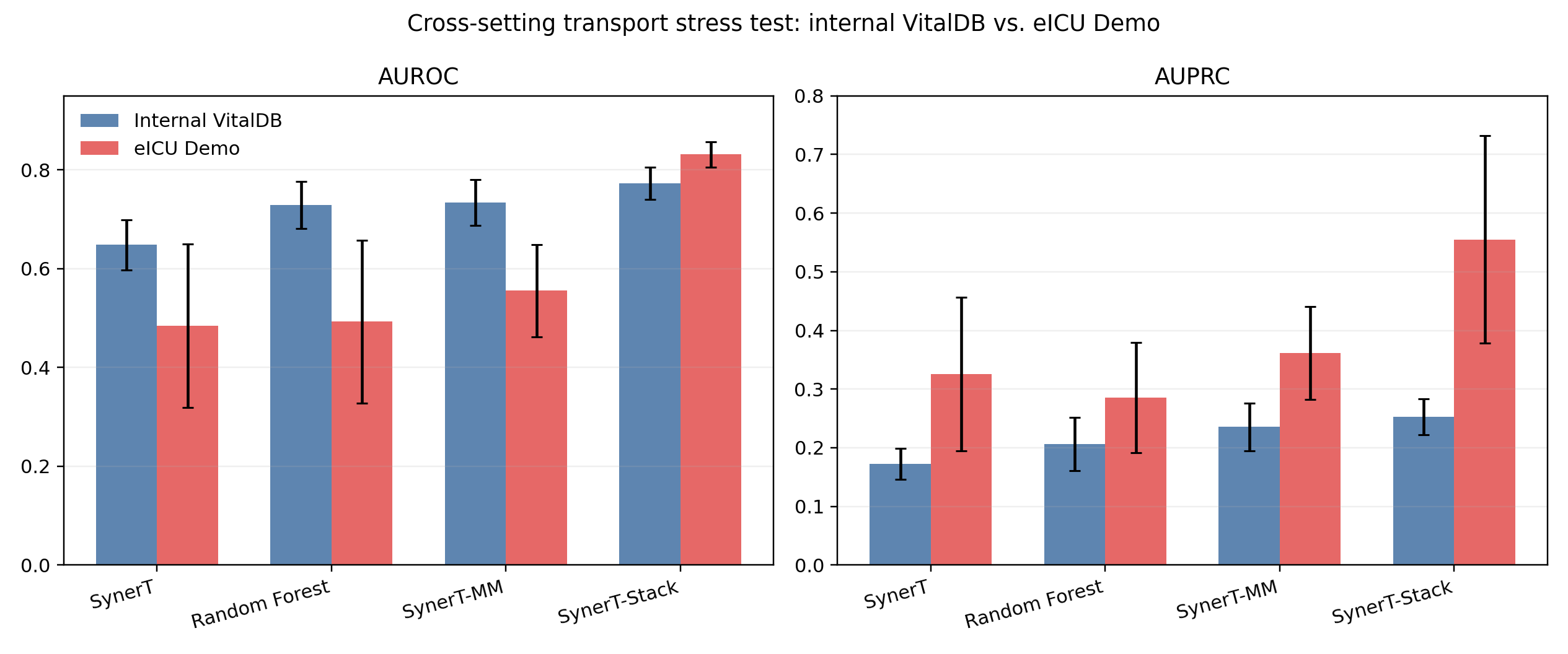}%
}{%
\missingfigurebox{0.8\textwidth}%
}
\caption{Cross-setting transport stress test comparing internal VitalDB and eICU Demo performance.}
\label{fig:transport_shift}
\end{figure}

\subsection{Statistical Testing of the Main Contribution}
\label{subsec:stat_testing}

\hyperref[tab:stat_tests]{Table~\ref*{tab:stat_tests}} reports the pre-specified paired AUPRC comparisons, using fold-paired Wilcoxon signed-rank tests together with paired bootstrap deltas from aligned out-of-fold predictions. Because random forest was the strongest AUROC baseline and one of the base learners in SynerT-Stack, the comparison path was anchored on the random-forest baseline. The results support three conclusions: random forest significantly outperformed the original waveform-only SynerT; SynerT-MM improved over random forest in point estimate but not at a statistically significant level in the current five-fold evaluation; and SynerT-Stack significantly improved over both SynerT-MM and random forest, making stacking the most statistically secure gain in the main analysis.

\begin{table}[t]
\caption{Paired statistical testing for the main contribution-focused AUPRC comparisons. Fold-level significance uses a Wilcoxon signed-rank test with one-sided alternative \emph{model A} $>$ \emph{model B}.}
\label{tab:stat_tests}
\centering
\small
\begin{tabular}{p{4.2cm}ccp{4.6cm}}
\hline\noalign{\smallskip}
Comparison & Mean $\Delta$ AUPRC & Wilcoxon $p$ & Paired bootstrap $\Delta$ AUPRC (95\% CI) \\
\noalign{\smallskip}\hline\noalign{\smallskip}
Random-forest baseline $>$ Original SynerT & +0.053 & 0.0313 & +0.057 \,[0.009, 0.108] \\
SynerT-MM $>$ Random-forest baseline & +0.019 & 0.0625 & +0.010 \,[{-}0.027, 0.047] \\
SynerT-Stack $>$ SynerT-MM & +0.027 & 0.0313 & +0.038 \,[0.003, 0.071] \\
SynerT-Stack $>$ Random-forest baseline & +0.046 & 0.0313 & +0.048 \,[0.015, 0.080] \\
\noalign{\smallskip}\hline\noalign{\smallskip}
\end{tabular}
\end{table}

\subsection{Calibration Assessment}
\label{subsec:calibration}

\hyperref[tab:calibration_results]{Table~\ref*{tab:calibration_results}} summarizes pooled out-of-fold calibration diagnostics for the main contribution path. Among the raw models, random forest showed the best calibration, whereas both SynerT-MM and SynerT-Stack were substantially miscalibrated. Cross-fitted Platt recalibration largely corrected this defect, reducing ECE from 0.350 to 0.005 for SynerT-MM and from 0.316 to 0.027 for SynerT-Stack while leaving discrimination largely unchanged. Thus, SynerT-Stack remained the best discriminator, but clinically interpretable absolute-risk use should rely on post-hoc recalibration.

\begin{table}[H]
\caption{Calibration diagnostics on pooled out-of-fold predictions for the main contribution path. Lower Brier and ECE are better; calibration intercept $0$ and slope $1$ are ideal. Boldface marks the best value in each column.}
\label{tab:calibration_results}
\centering
\small
\begin{tabular}{p{4.0cm}cccc}
\hline\noalign{\smallskip}
Model variant & Brier & ECE & Cal. intercept & Cal. slope \\
\noalign{\smallskip}\hline\noalign{\smallskip}
Random-forest baseline & 0.076 & 0.090 & -0.989 & \textbf{0.985} \\
SynerT-MM & 0.210 & 0.350 & -2.513 & 0.775 \\
SynerT-MM + Platt & \textbf{0.065} & \textbf{0.005} & \textbf{-0.104} & 0.956 \\
SynerT-Stack & 0.179 & 0.316 & -2.436 & 0.647 \\
SynerT-Stack + Platt & 0.065 & 0.027 & -0.143 & 0.941 \\
\noalign{\smallskip}\hline\noalign{\smallskip}
\end{tabular}
\end{table}

\subsection{Decision-Curve Analysis}
\label{subsec:dca}

\begin{figure}[H]
\sidecaption[t]
\Description{Decision-curve analysis comparing random forest, recalibrated SynerT-MM, recalibrated SynerT-Stack, and treat-all or treat-none strategies across threshold probabilities from 0.02 to 0.25.}
\IfFileExists{figure/revision_internal_dca.png}{%
\includegraphics[width=7.6cm]{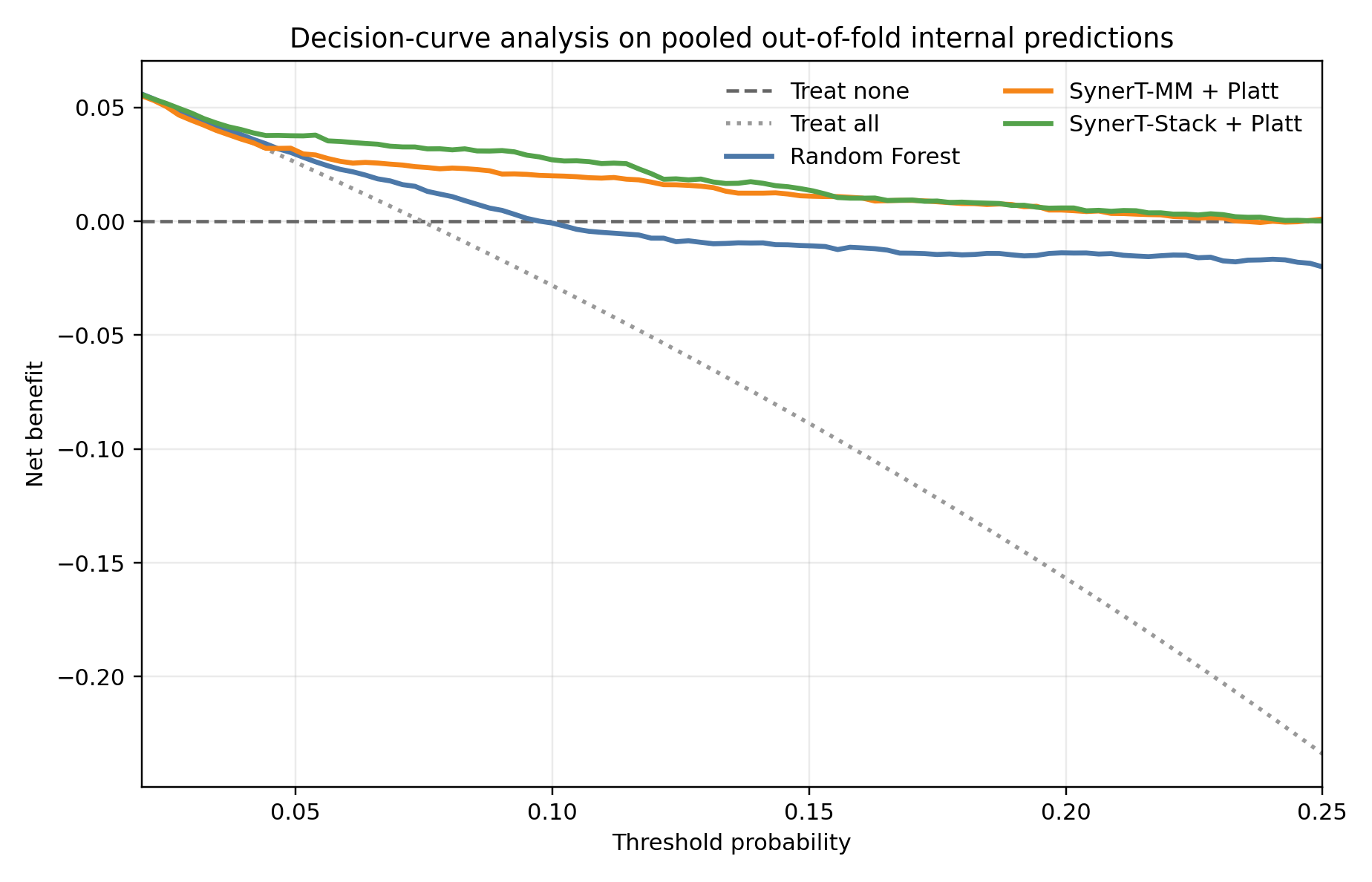}%
}{%
\missingfigurebox{7.6cm}%
}
\caption{Decision-curve analysis comparing random forest, recalibrated SynerT-MM, recalibrated SynerT-Stack, and treat-all or treat-none strategies across threshold probabilities from 0.02 to 0.25.}
\label{fig:dca}
\end{figure}

\hyperref[fig:dca]{Figure~\ref*{fig:dca}} shows decision-curve analysis for the deployment-ready probabilities: raw random-forest scores and cross-fitted Platt-recalibrated SynerT-MM and SynerT-Stack probabilities. Both recalibrated multimodal models maintained positive net benefit across the examined threshold range of 0.02 to 0.25, whereas the random-forest baseline fell below the treat-none strategy at approximately the 0.10 threshold. Across most low-to-intermediate thresholds, the recalibrated SynerT-Stack model provided the highest net benefit, supporting it as the most defensible candidate for risk-guided use after leakage-safe recalibration.

\subsection{ROC, Precision-Recall, and Operating-Point Evaluation}
\label{subsec:curves}

\begin{figure}[t]
\sidecaption
\Description{Out-of-fold ROC curve panel for the principal model families.}
\IfFileExists{figure/main_results_roc_overlay.png}{%
\includegraphics[width=0.55\textwidth]{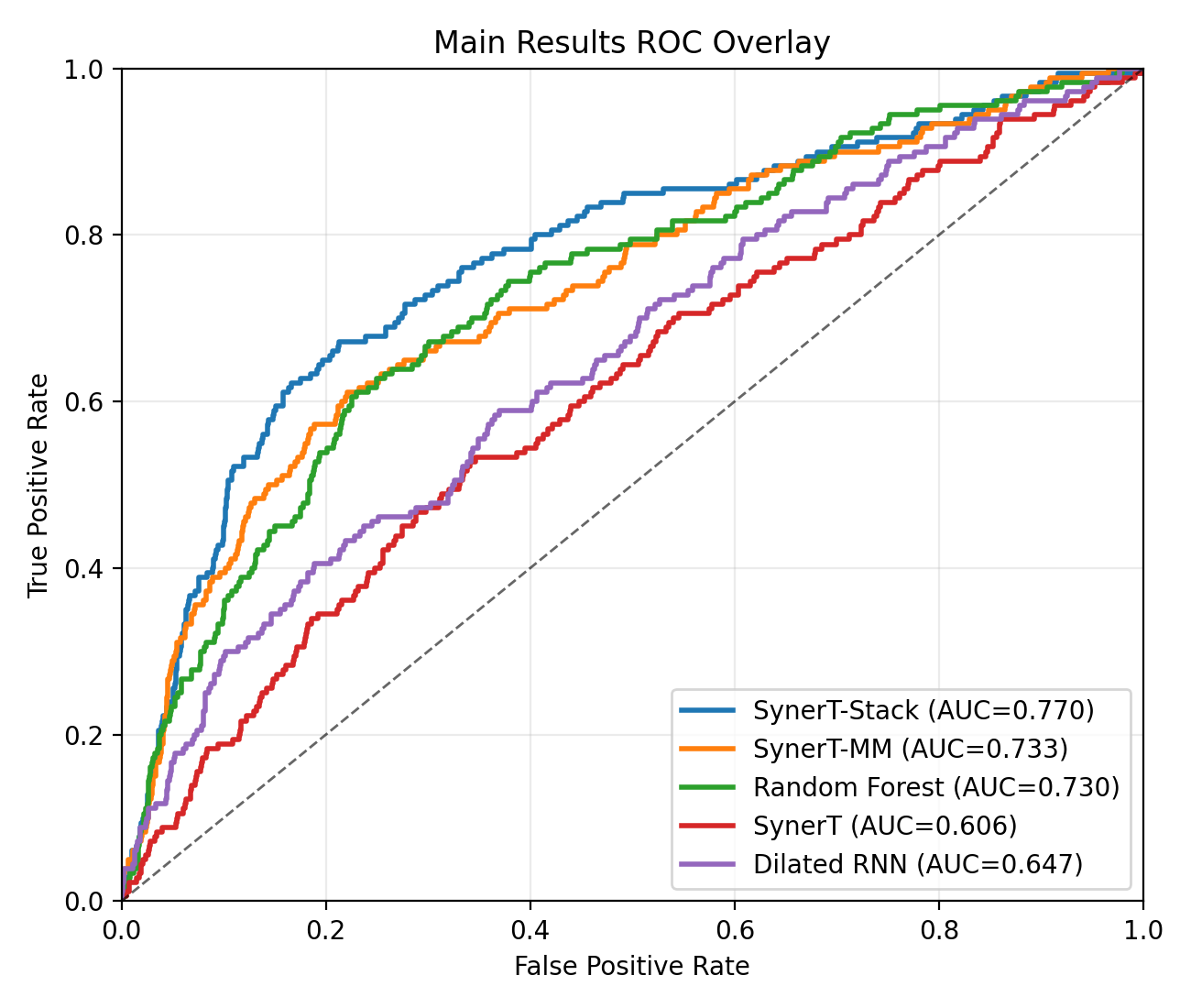}%
}{%
\missingfigurebox{0.55\textwidth}%
}
\caption{Out-of-fold ROC curves for the principal model families. SynerT-Stack provides the strongest threshold-agnostic discrimination, with SynerT-MM consistently above the waveform-only SynerT model.}
\label{fig:roc_overlay}
\end{figure}

\hyperref[fig:roc_overlay]{Figure~\ref*{fig:roc_overlay}} provides threshold-dependent ROC discrimination views for the principal model families. \hyperref[tab:operating_points]{Table~\ref*{tab:operating_points}} summarizes clinically oriented operating-point performance at 95\% specificity. At this operating point, SynerT-Stack achieved the highest sensitivity and positive predictive value, with SynerT-MM ranking next among the neural variants.

\begin{table}[t]
\caption{Clinically oriented operating-point metrics for the leading models and strongest tabular baselines. Values are mean $\pm$ standard deviation across five folds. Boldface marks the best value in each column.}
\label{tab:operating_points}
\centering
\small
\begin{tabular}{p{3.6cm}ccc}
\hline\noalign{\smallskip}
Model & Sens@Spec=0.95 & PPV@Spec=0.95 & F1-max \\
\noalign{\smallskip}\hline\noalign{\smallskip}
Original SynerT & 0.178 $\pm$ 0.058 & 0.252 $\pm$ 0.060 & 0.245 $\pm$ 0.042 \\
Random-forest baseline & 0.233 $\pm$ 0.032 & 0.298 $\pm$ 0.032 & 0.311 $\pm$ 0.041 \\
Extra-trees baseline & 0.167 $\pm$ 0.056 & 0.276 $\pm$ 0.110 & 0.307 $\pm$ 0.045 \\
CatBoost baseline & 0.222 $\pm$ 0.071 & 0.274 $\pm$ 0.057 & 0.315 $\pm$ 0.051 \\
SynerT-MM & 0.250 $\pm$ 0.094 & 0.311 $\pm$ 0.094 & 0.349 $\pm$ 0.050 \\
SynerT-Stack & \textbf{0.283} $\pm$ \textbf{0.041} & \textbf{0.328} $\pm$ \textbf{0.024} & \textbf{0.367} $\pm$ \textbf{0.034} \\
\noalign{\smallskip}\hline\noalign{\smallskip}
\end{tabular}
\end{table}

\section{Discussion}
\label{sec:discussion}

This study demonstrates that strict leakage-aware evaluation is essential for
credible early PO-AKI prediction, because without such constraints the apparent
performance of waveform-only models can be overstated. Under controlled early
prediction settings, waveform-only temporal modeling was insufficient for
reliable risk stratification, whereas integrating physiologic trajectories with
structured renal-risk context and hemodynamic burden summaries through
leakage-safe stacking produced the strongest overall results, supporting a
multimodal information-fusion view of the task rather than a purely architectural
one. The eICU-CRD Demo analysis should be interpreted as a transport stress test
rather than definitive external validation given its small, ICU-based, and
mismatched cohort \cite{pollard2018eicu,riley2024external,riley2021valsize}, yet
it still suggests that the stacked ensemble is less brittle under setting shift
than standalone models. At the same time, improved discrimination alone was not
sufficient for deployment-ready risk estimation, because the multimodal models
required leakage-safe Platt recalibration to correct important calibration
defects, after which decision-curve analysis showed superior net benefit for the
stacked model across low-to-intermediate thresholds. Clinically, prediction
within the first 60 intraoperative minutes may support earlier hemodynamic
optimization and ICU triage, but the present study remains limited by
creatinine-only AKI labeling without urine-output ascertainment
\cite{kdigo2012} and by the small unmatched external cohort, so validation on
larger perioperative datasets remains necessary. Overall, these findings suggest
that credible early PO-AKI prediction depends more on multimodal context,
calibration, and robustness assessment than on temporal-model complexity alone.
\section{Conclusion}
\label{sec:conclusion}

In this leakage-aware intraoperative prediction study, waveform-only temporal modeling was insufficient for postoperative AKI discrimination. Performance improved when dynamic vital-sign trajectories were combined with structured renal-risk context and clinically motivated hemodynamic burden summaries, and improved further through leakage-safe stacking of complementary learners. The strongest supported result is therefore the advantage of the stacked model, not a generic claim of waveform-only deep-learning superiority. Calibration and decision-curve analysis refined that conclusion: the stacked model delivered the best ranking performance, cross-fitted Platt recalibration improved the absolute-risk behavior of both multimodal models, and the recalibrated stacked model provided the most favorable clinical-utility profile across low-to-intermediate thresholds. The eICU-CRD Demo analysis adds a stress test, showing that this advantage is less brittle under cross-setting shift while motivating matched surgical cohorts for external validation. More broadly, the leakage-aware multimodal evaluation framework presented here offers a transferable template for early intraoperative risk stratification studies that disentangle the contributions of dynamic waveform data, structured clinical context, and calibration-aware ensemble design under strict clinical deployment constraints.


\clearpage
\begin{thebibliography}{99.}

\bibitem{kdigo2012} KDIGO Acute Kidney Injury Work Group (2012) KDIGO clinical practice guideline for acute kidney injury. Kidney Int Suppl 2(1):1--138. Available at: \url{https://kdigo.org/wp-content/uploads/2016/10/KDIGO-2012-AKI-Guideline-English.pdf}

\bibitem{lee2022vitaldb} Lee HC, Park Y, Yoon SB, Yang SM, Park D, Jung CW (2022) VitalDB, a high-fidelity multi-parameter vital signs database in surgical patients. Sci Data 9(1):279. \url{https://doi.org/10.1038/s41597-022-01411-5}

\bibitem{pollard2018eicu} Pollard TJ, Johnson AEW, Raffa JD, Celi LA, Mark RG, Badawi O (2018) The eICU Collaborative Research Database, a freely available multi-center database for critical care research. Sci Data 5:180178. \url{https://doi.org/10.1038/sdata.2018.178}

\bibitem{peng2021aki}
Peng YC, D'Souza NS, Bush B, Brown C, Venkataraman A (2021)
Predicting acute kidney injury via interpretable ensemble learning and attention weighted convoutional-recurrent neural networks.
In: 2021 55th Annual Conference on Information Sciences and Systems (CISS), Baltimore, MD, USA, pp 1--6. IEEE.
\url{https://doi.org/10.1109/CISS50987.2021.9400242}

\bibitem{prowle2021poaki} Prowle JR, Forni LG, Bell M, et al. (2021) Postoperative acute kidney injury in adult non-cardiac surgery: joint consensus report of the Acute Disease Quality Initiative and Peri-Operative Quality Initiative. Nat Rev Nephrol 17(9):605--618. \url{https://doi.org/10.1038/s41581-021-00418-2}

\bibitem{kline2022multimodal} Kline A, Wang H, Li Y, et al. (2022) Multimodal machine learning in precision health: a scoping review. npj Digit Med 5(1):171. \url{https://doi.org/10.1038/s41746-022-00712-8}

\bibitem{zhang2022aki} Zhang H, Wang Y, Wu S, et al. (2022) Artificial intelligence for the prediction of acute kidney injury during the perioperative period: systematic review and meta-analysis of diagnostic test accuracy. BMC Nephrol 23(1):405. \url{https://doi.org/10.1186/s12882-022-03025-w}

\bibitem{park2025aki} Park S, Chung S, Kim Y, et al. (2025) A deep-learning algorithm using real-time collected intraoperative vital sign signals for predicting acute kidney injury after major non-cardiac surgeries: a modelling study. PLoS Med 22(4):e1004566. \url{https://doi.org/10.1371/journal.pmed.1004566}

\bibitem{lee2024open} Lee SW, Jang J, Seo WY, Lee D, Kim SH (2024) Internal and external validation of machine learning models for predicting acute kidney injury following non-cardiac surgery using open datasets. J Pers Med 14(6):587. \url{https://doi.org/10.3390/jpm14060587}

\bibitem{vancalster2019} Van Calster B, McLernon DJ, van Smeden M, Wynants L, Steyerberg EW, Topic Group `Evaluating diagnostic tests and prediction models' of the STRATOS initiative (2019) Calibration: the Achilles heel of predictive analytics. BMC Med 17(1):230. \url{https://doi.org/10.1186/s12916-019-1466-7}

\bibitem{collins2024tripodai} Collins GS, Moons KGM, Dhiman P, et al. (2024) TRIPOD+AI statement: updated guidance for reporting clinical prediction models that use regression or machine learning methods. BMJ 385:e078378. \url{https://doi.org/10.1136/bmj-2023-078378}

\bibitem{riley2024external} Riley RD, Archer L, Snell KIE, Ensor J, Dhiman P, Martin GP, Bonnett LJ, Collins GS (2024) Evaluation of clinical prediction models (part 2): how to undertake an external validation study. BMJ 384:e074820. \url{https://doi.org/10.1136/bmj-2023-074820}

\bibitem{riley2021valsize} Riley RD, Debray TPA, Collins GS, Archer L, Ensor J, van Smeden M, Snell KIE (2021) Minimum sample size for external validation of a clinical prediction model with a binary outcome. Stat Med 40(19):4230--4251. \url{https://doi.org/10.1002/sim.9025}

\bibitem{vickers2006dca} Vickers AJ, Elkin EB (2006) Decision curve analysis: a novel method for evaluating prediction models. Med Decis Making 26(6):565--574. \url{https://doi.org/10.1177/0272989X06295361}

\bibitem{vickers2021dca} Vickers AJ, Holland F (2021) Decision curve analysis to evaluate the clinical benefit of prediction models. Spine J 21(10):1643--1648. \url{https://doi.org/10.1016/j.spinee.2021.02.024}

\end{thebibliography}
\end{document}